\documentclass[letterpaper]{article} 
\usepackage{aaai25}  
\usepackage{times}  
\usepackage{helvet}  
\usepackage{courier}  
\usepackage[hyphens]{url}  
\usepackage{graphicx} 
\usepackage{natbib}  
\usepackage{caption} 
\usepackage{algorithm}
\usepackage{algorithmic}
\usepackage{newfloat}
\usepackage{listings}
\DeclareCaptionStyle{ruled}{labelfont=normalfont,labelsep=colon,strut=off} 
\floatstyle{ruled}
\newfloat{listing}{tb}{lst}{}
\floatname{listing}{Listing}
\usepackage{ragged2e} 
\usepackage{booktabs,makecell, multirow, tabularx}
\usepackage{array}
\nocopyright

\title{DA-STGCN: 4D Trajectory Prediction Based on Spatiotemporal Feature Extraction}
\author{
    Yuheng Kuang\textsuperscript{\rm 1}, Zhengning Wang\textsuperscript{\rm 1}, Jianping Zhang\textsuperscript{\rm 2}, Zhenyu Shi\textsuperscript{\rm 1}, Yuding Zhang\textsuperscript{\rm 1}
}
\affiliations{
    \textsuperscript{\rm 1}School of Information and Communication Engineering, University of Electronic Science and Technology of China\\
    \textsuperscript{\rm 2}Second Research Institute of Civil Aviation Administration of China

}

\usepackage{bibentry}

\begin{document}

\maketitle

\begin{abstract}
The importance of four-dimensional (4D) trajectory prediction within air traffic management systems is on the rise. Key operations such as conflict detection and resolution, 
aircraft anomaly monitoring, and the management of congested flight paths are increasingly reliant on this foundational technology, underscoring the urgent demand for 
intelligent solutions. The dynamics in airport terminal zones and crowded airspaces are intricate and ever-changing; however, current methodologies do not sufficiently 
account for the interactions among aircraft. To tackle these challenges, we propose DA-STGCN, an innovative spatiotemporal graph convolutional network that integrates a 
dual attention mechanism. Our model reconstructs the adjacency matrix through a self-attention approach, enhancing the capture of node correlations, and employs 
graph attention to distill spatiotemporal characteristics, thereby generating a probabilistic distribution of predicted trajectories. 
This novel adjacency matrix, reconstructed with the self-attention mechanism, is dynamically optimized throughout the network's training process, offering a more nuanced 
reflection of the inter-node relationships compared to traditional algorithms. The performance of the model is validated on two ADS-B datasets, one near the airport 
terminal area and the other in dense airspace. Experimental results demonstrate a notable improvement over current 4D trajectory prediction methods, 
achieving a 20\% and 30\% reduction in the Average Displacement Error (ADE) and Final Displacement Error (FDE), respectively. 
The incorporation of a Dual-Attention module has been shown to significantly enhance the extraction of node correlations, as verified by ablation experiments.
\end{abstract}

%

\section{Introduction}

The International Civil Aviation Organization (ICAO) released the 6th edition of the Global Air Navigation Plan in 2019. It explicitly states that 
Trajectory-Based Operations (TBO) will serve as the overarching integration and ultimate implementation goal for various elements within the 
``Air Traffic Management (ATM) Systems Block Upgrade'', with plans to deploy after 2031 globally. The transition of Air Traffic Management (ATM) from 
``Command-Based Operations'' (CBO) to ``Trajectory-Based Operations'' (TBO) brings new challenges to the functionality and performance of 4D trajectory prediction \cite{ruiz2018novel, song2012improved}.

As the core technology for different tasks such as conflict detection and resolution, aircraft anomaly behavior monitoring, dense traffic area management, 
and arrival and departure sequencing, 4D trajectory prediction has to face one of the main challenges in current research on how to accurately 
and efficiently predict future trajectories.The US Federal Aviation Administration (FAA) and European Control Action Plan 16 (AP16) defines the 4D trajectory as a four-dimensional description of an 
aircraft's flight path: latitude, longitude, altitude, and time \cite{zeng2022aircraft}. The International Civil Aviation Organization (ICAO) extends this definition to ground operations, 
representing a trajectory as the description of an aircraft's movement both in the air and on the ground, including attributes such as position, time, speed, 
and acceleration \cite{organisation2005global}. 
\begin{figure}[t]
\centering
\includegraphics[width=0.9\columnwidth]{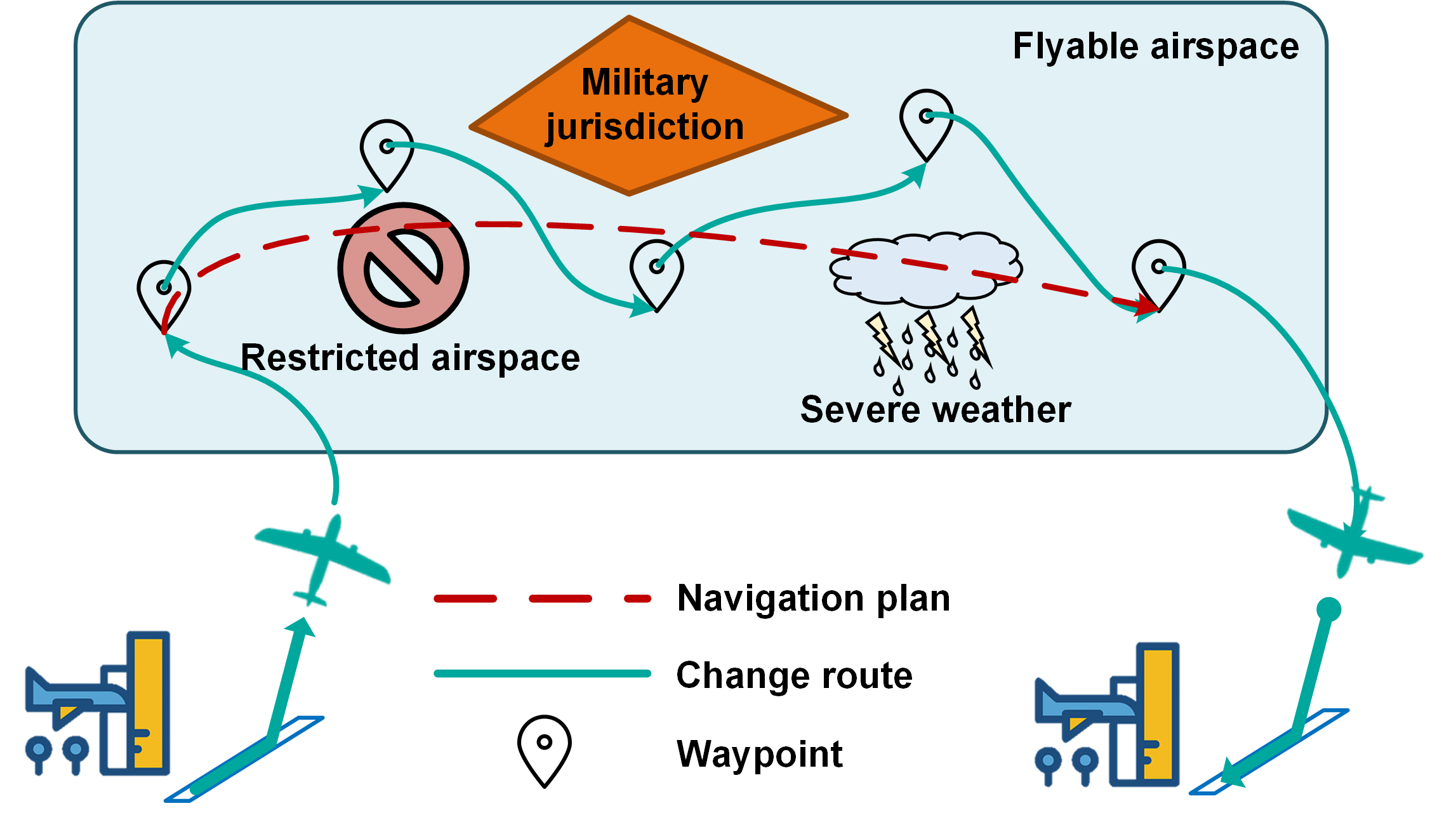} 
\caption{The actual flight path of aircraft sometimes doesn't follow the pre-planned route. Advanced 4D trajectory prediction can assist ATM in better handling these emergencies.}
\label{fig1}
\end{figure}

Due to unpredictable factors such as crowded airspace, adverse weather, temporary military activities, and emergency incidents, 
the actual flight path of an aircraft sometimes cannot follow the pre-planned route, as shown in Figure 1. These unexpected situations reduce the predictability of 
air traffic, increase the difficulty of ATM, and severely hinder the efficient operation of ATM systems. If the future trajectory of an aircraft could be 
predicted in advance, it would be beneficial for managers to make proactive decisions to handle unexpected situations, thereby ensuring flight safety and efficiency. 
Therefore, it is necessary to predict the 4D trajectory of the aircraft.

\section{Related works}
In recent years, research on 4D trajectory prediction methods can be generally classified into three categories: state estimation methods, dynamics-based methods, 
and machine learning methods. The state estimation models establish motion equations according to the aircraft's position, velocity, acceleration, 
and other attributes to achieve estimation propagation. This process does not involve the relationships between aircraft mass, acceleration, angle, and other states, 
making the model relatively simple. However, due to the inability to accurately quantify the long-term maneuver uncertainties of aircraft, it results in significant 
errors and is only effective for short-term predictions. Typical estimation methods include Kalman filtering 
algorithms \cite{avanzini2004frenet, wang20144d}, particle filtering algorithms \cite{lymperopoulos2010sequential}, and hidden Markov models \cite{ayhan2016aircraft, lin2019approach}. 

Dynamics-based trajectory prediction models can be regarded as a kind of physical model, which predict the continuous future trajectory points of an 
aircraft based on the forces acting on it. Specifically, given the current state of the aircraft, meteorological conditions, aircraft performance parameters, 
and flight intentions, the aircraft dynamics model is built using differential equations, and future continuous trajectory points are predicted by solving these equations. 
The most commonly used dynamics model is the point mass model \cite{weitz2015derivation, lee2016hybrid, chang20234d}. Although the dynamics model analyzes from the perspective of the force on the aircraft, 
in order to simplify the model, most studies are conducted under ideal assumptions, rarely considering actual constraints, 
resulting in predictions that do not meet real-world needs.

Currently, more research is focused on machine learning methods. Traditional machine learning algorithms, such as regression models \cite{de2013machine, tastambekov2014aircraft, kanneganti2018visualization}, 
clustering algorithms \cite{georgiou2020semantic, wang2017short, barratt2018learning}, and etc., have surpassed past state estimation models and dynamics-based methods in prediction accuracy 
but still struggle to achieve satisfactory performance in real-time predictions. Subsequently, more algorithms based on deep neural networks have been proposed, 
with recurrent neural networks (RNNs) and their improved models being widely used due to their ability to mine time series information\cite{shi20204, ma2020hybrid, zeng2020deep, ma2024data}. 

\cite{xu2021multi} introduced the Social-LSTM method into aircraft trajectory prediction. Previously, Social-LSTM was mainly used for pedestrian trajectory prediction tasks \cite{alahi2016social}. 
In Social-LSTM, future trajectory sequences and arrival times are predicted not only based on historical trajectories but also by 
considering the trajectories of other nearby aircraft. Compared with traditional LSTM methods, Social-LSTM reduces the data dimensions required 
for trajectory prediction and provides more accurate prediction results. 
However, the use of recurrent architectures has limitations; it struggles to effectively extract the implicit interactions between aircraft and has low training efficiency.

To address the limitations of recurrent architectures, some studies have adopted graph convolutional neural networks (GCNs) for trajectory prediction. 
One study \cite{fan2024global} combined LSTM and GCN to propose a graph network model based on global and local attribute relationships, 
achieving good results in solving aircraft trajectory prediction problems. Another study \cite{yan2018spatial} proposed a spatiotemporal graph convolutional neural network (STGCNN) 
to predict human dynamic behavior. Compared to traditional GCNs, STGCNN can simultaneously capture the temporal and spatial features of nodes, 
demonstrating better performance in time-series prediction tasks. \cite{mohamed2020social} was the first to apply STGCNN to pedestrian trajectory prediction tasks, 
proposing a Social-STGCNN model. Social-STGCNN fully captures the spatiotemporal correlations between nodes by constructing the relationships 
between edges of nodes in a social graph. It uses temporal extrapolation convolutional networks (TXP-CNN) to predict the distribution of trajectories over time. 
Subsequently, STGCNN has been used for aircraft 4D trajectory prediction and ship trajectory prediction tasks \cite{xu2023aircraft, liao2024dynamic}, achieving good results.

Although STGCNN has achieved good results in trajectory prediction tasks, these methods typically employ artificially defined kernel functions to 
calculate adjacency matrices that describe the relationships between nodes. In trajectory prediction tasks, this kernel function is usually 
defined as the inverse of the distance between nodes \cite{mohamed2020social}. This adjacency matrix obtained using only distance relationships cannot fully reflect 
the real potential relationships between aircraft in highly dynamic and complex scenarios such as airport terminal areas or dense airspace. 
Additionally, GCNs cannot handle dynamic graph problems and are not easy to implement different learning weights to different neighbor nodes, 
leading to the inability to effectively aggregate the correlations between nodes. This paper addresses the above issues by proposing a 
dual-attention based spatiotemporal graph convolutional network (DA-STGCN) model, achieving better results in aircraft trajectory prediction tasks compared to 
existing methods. The main contributions are as follows:

1) We propose a learnable adjacency matrix based on self-attention mechanism. First, the initial adjacency matrix is constructed using the 
inverse distance between aircrafts. Then, the self-attention mechanism is used to reconstruct and continuously optimize the adjacency matrix during network 
training to improve its generalization. This effectively solves the problem where distance-based adjacency matrices cannot fully reflect the potential 
relationships between aircraft in dynamic scenarios.

2) We integrate graph attention network (GAT) to further aggregate the correlations between nodes. The problem of GCN not being able to effectively aggregate 
the relevance between nodes using the GAT structure was settled, leading to the proposal of our model DA-STGCN.
\begin{figure*}[t]
\centering
\includegraphics[width=0.85\textwidth]{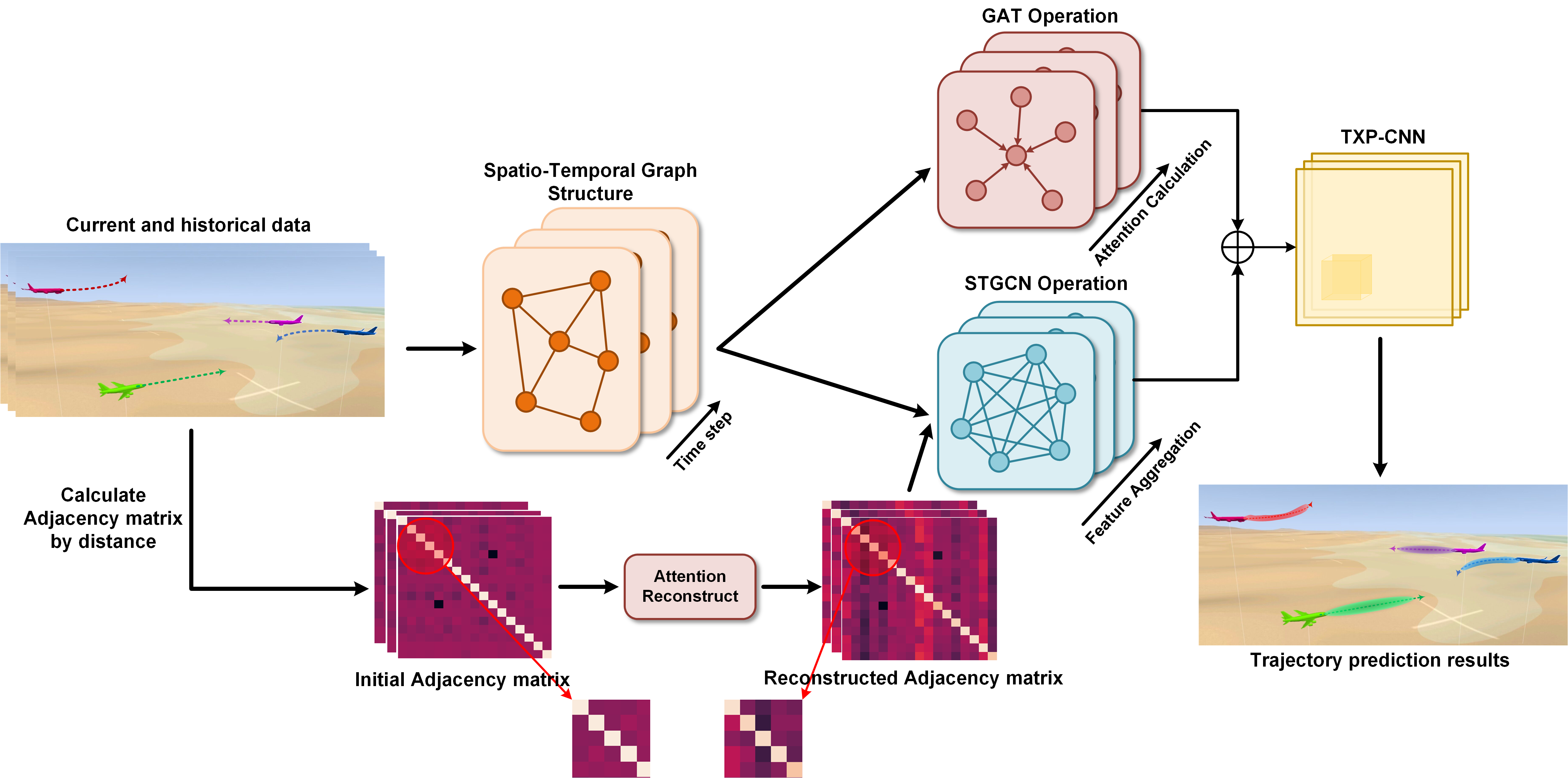} 
\caption{Overall framework of DA-STGCN model.}
\label{fig2}
\end{figure*}

3) We experimentally validate our model on two datasets based on real world ADS-B data, one near the airport terminal area and the other in dense airspace. 
The results of comparison and ablation experiments verified that our proposed model can effectively capture the potential connections between aircraft and 
fully aggregate the correlations between nodes.

\section{Methodology}
\subsection{The DA-STGCN model}
Figure 2 illustrates the framework of the DA-STGCN we proposed. The dual-attention module consists of two components. 
First, we utilize a self-attention mechanism to reconstruct the adjacency matrix, which was initially defined by the reciprocal of the distance between nodes. 
Then, we perform spatiotemporal graph convolution operations with the reconstructed adjacency matrix and the spatiotemporal graph. 
This process constitutes the first part of the dual-attention module. The second part of the dual-attention module employs GAT to calculate 
attention on the initial spatiotemporal graph, yielding the second part of the output, which is beneficial for further aggregating the correlations 
between nodes on the spatiotemporal graph. It is worth noting that the GAT does not rely on the adjacency matrix. Consequently, the two components 
of our dual-attention module operate independently, enabling each to extract distinct node relationship features. This independence allows us to 
ultimately converge these features, thereby making more comprehensive use of the potential correlations between nodes. 
Finally, we converge these features and input them into the TXP-CNN to obtain the results of the 4D trajectory prediction for aircraft. 

\subsection{Construction of spatiotemporal graph}
Graph convolution, as a core operation in graph neural networks, is extensively used for processing data with graph structures. 
The purpose of graph convolution computation is to fully utilize the information of nodes and edges within the graph structure. 
By performing graph convolution operations on nodes and adjacency matrices, we can extract features for each node.

We first construct a spatiotemporal graph based on aircraft trajectory data, which is necessary for graph convolution operations. 
We define a spatial graph at a certain moment $t$ as $G_t=(V_t,E_t)$, which contains nodes and edges. 
$V_t=\{v_t^i\ |\forall i\in{1,\ldots N}\}$ is the collection of node vectors in $G_t$, and each element $v_t^i$ contains the features of 
the $i$-th aircraft at moment $t$. In our work, these features contain the longitude, latitude, and altitude of the aircraft, as shown in Eq. (1):
\begin{equation}
\label{equation1}
v_t^i=[x_t^i,y_t^i,z_t^i]
\end{equation}

$E_t=\{e_t^{ij}\ |\forall i,j\in\{1,\ldots N\}\}$ is the collection of all edges in $G_t$, and each element $e_t^{ij}$ represents the connection relationship 
between nodes $v_t^i$ and $v_t^j$, which can be seen as the mutual influence between two aircraft. We use a kernel function to measure the interaction 
between two aircraft, which is same as \cite{mohamed2020social}, and the kernel function $k(v_t^i,v_t^j)$ is defined as Eq. (2):
\begin{equation}
\label{equation2}
k(v_{t}^{i}, v_{t}^{j})=\left\{\begin{array}{ll}{1}/{\|v_{t}^{i}-v_{t}^{j}\|_{2}} & ,\|v_{t}^{i}-v_{t}^{j}\|_{2} \neq 0 \\0 & , {Otherwise}\end{array}\right.
\end{equation}

The physical meaning of $k(v_t^i,v_t^j)$ is the reciprocal of the distance between two aircraft, which means that the closer the distance between two aircraft, 
the greater their mutual influence, and vice versa. Therefore, we can construct the adjacency matrix as Eq. (3) shows:
\begin{equation}
\label{equation3}
A_{t}=\left[\begin{array}{cccc}
0 & e_{t}^{12} & \ldots & e_{t}^{1 N} \\
e_{t}^{21} & 0 & & e_{t}^{2 N} \\
& & \ddots & \vdots \\
e_{t}^{N 1} & e_{t}^{N 2} & \cdots & 0
\end{array}\right]
\end{equation}
where $A_t\in R^{N\times N}$ reflects the interrelationships of influence between all nodes in $G_t$.

For the convenience of subsequent graph convolution operations, the adjacency matrix $A_t$ needs to be normalized as shown in Eq. (4):
\begin{equation}
\label{equation4}
\bar{A}_{t}=\widehat{D}_{t}^{-\frac{1}{2}} \hat{A}_{t} \widehat{D}_{t}^{-\frac{1}{2}}
\end{equation}
where ${\hat{A}}_t=A_t+I$ and $I$ is the identity matrix, and ${\hat{D}}_t$ is the diagonal node degree matrix of the normalized adjacency matrix ${\bar{A}}_t$.

By stacking all $G_t$ at moment $t\in\{1\ldots T\}$, we can get the final spatiotemporal graph $G=(V,E)=\{G_t|\forall t\in\{1,\ldots T\}\}$, 
where $V\in R^{3\times T\times N}=\{V_t|\forall t\in\{1,\ldots T\}\}$. $E$ can be characterized by $\bar{A}\in R^{T\times N\times N}=\{{\bar{A}}_t|\forall t\in\{1,\ldots T\}\}$, 
where $N$ represents the number of aircraft. It is worth noting that all $G_t$ at moment $t\in\{1\ldots T\}$ must have the same topological structure. 
The feature vectors of each node in $V_t$ change only when $t$ changes.

\subsection{Reconstruct adjacency matrix based on attention}
As we mentioned earlier, the elements in the adjacency matrix defined by the kernel function $k(\cdot)$ can only reflect the distance between two aircraft 
and cannot reflect more potential associations between them. To solve this problem, considering the strong global information gathering capability of 
the attention mechanism, we propose an attention reconstruction module that uses self-attention to optimize the initial adjacency matrix. 
The specific process is shown in Figure 3.
\begin{figure}[t]
\centering
\includegraphics[width=0.95\columnwidth]{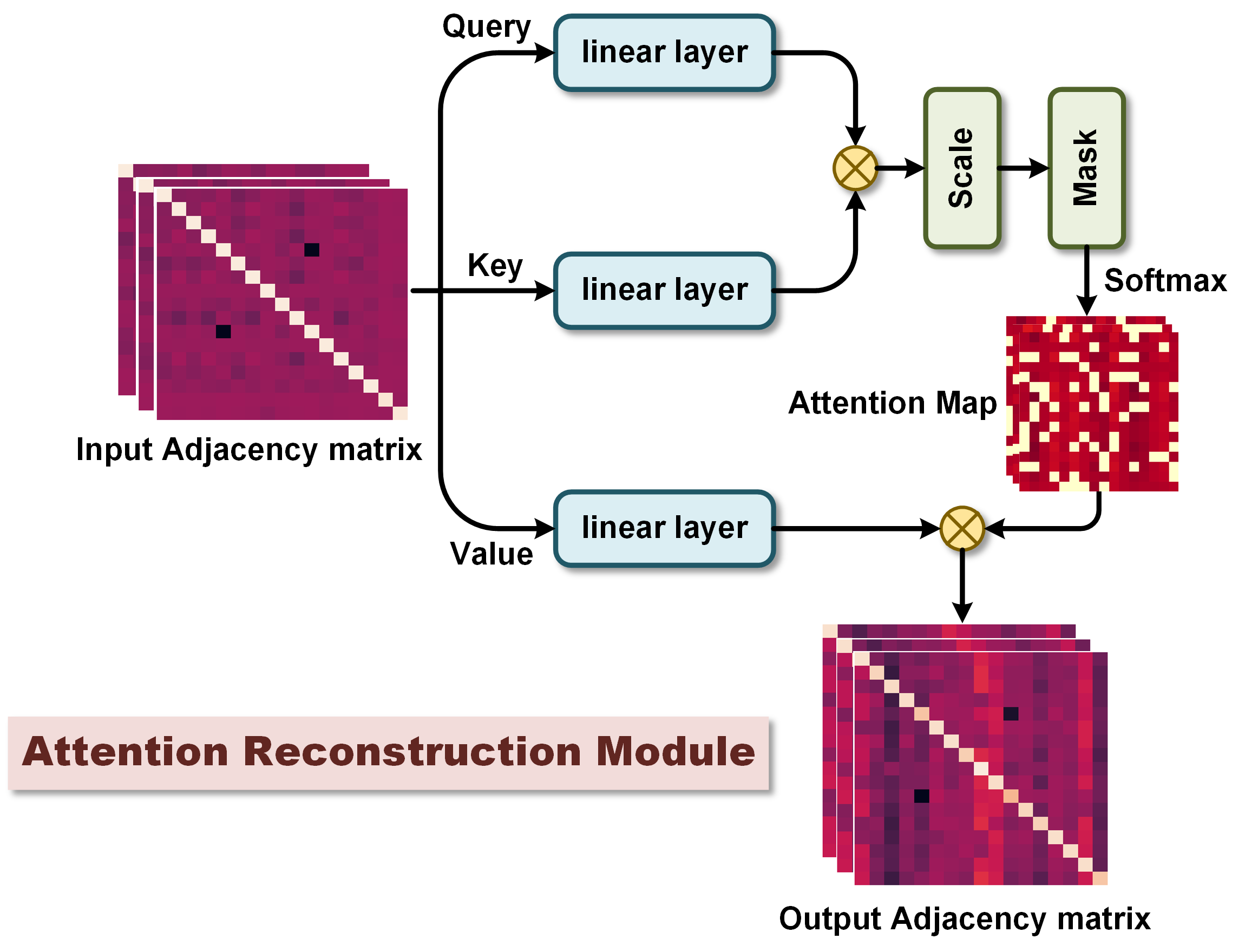} 
\caption{Adjacency Matrix Reconstruction Module Based on Self-Attention Mechanism.}
\label{fig3}
\end{figure}

We first use three different trainable weight matrices $\mathbf{W}_{q}, \mathbf{W}_{k}$ and $\mathbf{W}_{v}$ to perform linear 
transformations on the input adjacency matrix, generating corresponding feature vectors as Eq. (5) shows:
\begin{equation}
\label{equation5}
q_{i}=\mathbf{W}_{q} \bar{A}_{i}, \quad k_{j}=\mathbf{W}_{k} \bar{A}_{j}, \quad v_{j}=\mathbf{W}_{v} \bar{A}_{j}
\end{equation}
where ${\bar{A}}_i$ and ${\bar{A}}_j$ are the $i$-th and $j$-th rows of the adjacency matrix $\bar{A}$ respectively, 
which represent the connection relationship between nodes $i$ and $j$ in the graph.

Then, calculate the attention score and normalize it to obtain the attention weight $\alpha_{ij}$:
\begin{equation}
\label{equation6}
\begin{array}{c}
e_{i j}=q_{i}^{T} k_{j} \\ \\
\alpha_{i j}=\frac{\exp(e_{i j})}{\sum_{k=1}^{n} \exp(e_{i k})}
\end{array}
\end{equation}
where $q_i$ is the Query vector of node $i$, $k_j$ is the Key vector of node $j$, and $e_{ij}$ reflects the association between node $i$ and node $j$.

Ultimately, we sum the feature vectors $v_j$ with the attention weights $\alpha_{ij}$ to obtain the reconstructed features $\mathbf{h}_i^\prime$ of the nodes $i$:
\begin{equation}
\label{equation7}
\mathbf{h}_{i}^{\prime}=\sum_{j=1}^{n} \alpha_{i j} v_{j}
\end{equation}

We splice the features of all nodes to obtain the reconstructed adjacency matrix $A^\prime$. $A^\prime$ can dynamically adjust the relationship weights 
based on global information, not limited to predefined local connections. This makes the model able to better capture the implicit complex relationships 
between nodes, improving the predictive ability for the future trajectory of aircraft.

\subsection{Feature extraction by STGCN and GAT}
\textbf{STGCN operation.} We first use the reconstructed adjacency matrix $A^\prime$ for STGCN operation. We define the features of all aircraft at moment $t$ in 
the $l$-th network layer as $V_t^{(l)}$, and the definition of $V_t$ has been discussed in Section 3.2. Therefore, we use $V^{(l)}$ to represent the stacking 
of $V_t^{(l)}$ in the time dimension. Based on the above definition, we implement the graph convolution operation of each layer of STGCN with Eq. (8):
\begin{equation}
\label{equation8}
V^{(l+1)}=\sigma(A^{\prime} V^{(l)} \mathbf{W}^{(l)}_{s})
\end{equation}
where $\mathbf{W}^{l}_{s}$ is the trainable network parameter of the $l$-th layer, and $\sigma(\cdot)$ is the activation function. 
Finally, the output embedding of STGCN is represented as $\widetilde{V}\in R^{T\times\hat{M}\times N}$, where $\hat{M}$ is the dimension of output.

\textbf{GAT operation.} In GCN, node information aggregation relies on the fixed weights of the adjacency matrix within the graph, which can result in 
suboptimal information propagation. To address this, we employ GAT for enhanced feature extraction in spatiotemporal graphs. As depicted in Figure 4, 
this process runs concurrently with the operations of STGCN. Unlike GCN, GAT does not depend on a predefined adjacency matrix. 
Instead, it leverage self-attention mechanisms to dynamically allocate varying weights to each neighboring node. Thus, it allows the model to detect more 
nuanced relationships between nodes, transcending the constraints of proximity-based connections. 
We implement the calculation of the attention coefficients in GAT with Eq. (9) \cite{velickovic2017graph}:
\begin{equation}
\label{equation9}
\alpha_{i j}=\frac{\exp({LeakyReLU}(\mathbf{a}^{T}[\mathbf{W} \mathbf{h}_{i} \| \mathbf{W} \mathbf{h}_{\mathbf{j}}]))}{\sum_{k \in N_{i}} \exp({LeakyReLU}(\mathbf{a}^{T}[\mathbf{W} \mathbf{h}_{i} \| \mathbf{W}\mathbf{h}_{\mathbf{k}}]))}
\end{equation}
where $\alpha_{ij}$ is the attention coefficient from node $i$ to node $j$ and $N_i$ represents the set of neighbor nodes of node $i$. 
The input features of node are $\mathbf{h}=\{\mathbf{h}_1,\mathbf{h}_2,\ldots\mathbf{h}_N\}$ , where the number of nodes is $N$, and the 
output features of the nodes are $\mathbf{h}^\prime=\{{\mathbf{h}^\prime}_1,{\mathbf{h}^\prime}_2,\ldots{\mathbf{h}^\prime}_N\}$. $\mathbf{W}$ is the 
weight matrix of linear transformation applied to each node, and $\mathbf{a}$ is the weight vector that can map inputs to the feature space. 
Finally, LeakyReLU is used to provide non-linearity and SoftMax function for normalization. The final output feature of the node is:
\begin{equation}
\label{equation10}
\mathbf{h}_{i}^{\prime}=\|_{k=1}^{K} \sigma\left(\sum_{j \in N_{i}} \alpha_{i j}^{k} \mathbf{W}^{k} \mathbf{h}_{j}\right)
\end{equation}
where $\alpha_{ij}^k$ is the normalized attention coefficient of the $k$-th attention head, and $\|$ represents the concatenation operation.
\begin{figure}[t]
\centering
\includegraphics[width=0.95\columnwidth]{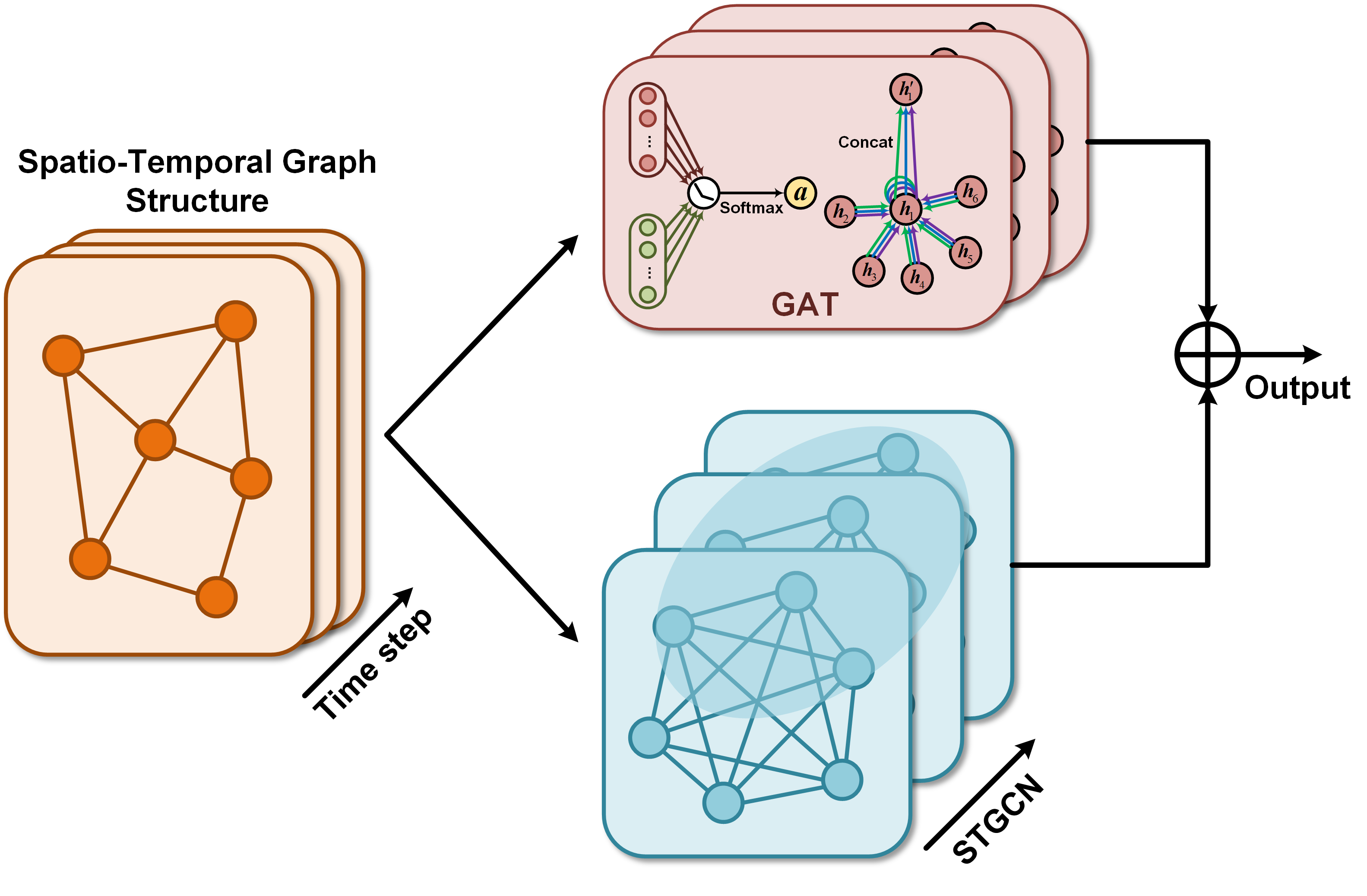} 
\caption{Operation STGCN and GAT.}
\label{fig4}
\end{figure}
Ultimately, we aggregate the outputs of STGCN and GAT to obtain results that include both potential features based on global distance relationships 
and additional features based on global node information.

\subsection{Trajectory prediction and loss design}
We implement the prediction of the aircraft's trajectory for continuous future time periods through TXP-CNN, which is consistent with \cite{mohamed2020social}. 
TXP-CNN takes the embedding results of STGCN and GAT as inputs and performs operations on the time dimension to achieve trajectory prediction. 
Since TXP-CNN performs convolution operations in the feature space, it is superior to recurrent architectures in terms of parameter volume and computational efficiency.

We use the probability distribution of the aircraft's future trajectory as the prediction result. We assume that the trajectory of the aircraft at 
future times satisfies a ternary Gaussian distribution, $(x_t^i,y_t^i,z_t^i)\sim \mathcal{N}(\mu_t^i,\Sigma_t^i)$. Therefore, the output of TXP-CNN should be 
the mean $\mu_t^i={(\mu_x,\mu_y,\mu_z)}_t^i$ and the elements in the covariance matrix $\Sigma_t^i$ that can determine a specific ternary Gaussian distribution, 
where $t$ and $i$ represent the predicted time and the aircraft number respectively. However, the output of the network cannot always satisfy the positive 
definiteness of the covariance matrix, so we output a lower triangular matrix $L_t^i\in R^{3\times3}$ with positive diagonal elements and construct the 
covariance matrix $\Sigma_t^i$ with $L_t^i$:
\begin{equation}
\label{equation11}
\Sigma_{t}^{i}=L_{t}^{i} \cdot\left(L_{t}^{i}\right)^{T}
\end{equation}

Finally, our model is trained to minimize the negative log-likelihood loss, which is defined as:
\begin{equation}
\label{equation12}
L^{i}(\mathbf{W}_{loss})=-\sum_{t=1}^{T_{P}} \log \left(P\left(\left(x_{t}^{i}, y_{t}^{i}, z_{t}^{i}\right) \mid \mu_{t}^{i}, \Sigma_{t}^{i}\right)\right)
\end{equation}
where $\mathbf{W}_{loss}$ represents all learnable parameters of the model, and $T_P$ is the length of the predicted time series.

\section{Experiment and Evaluation}
\subsection{Datasets and evaluation metrics}
Currently, there is no publicly available dataset for 4D trajectory prediction of aircraft using ADS-B data. Therefore, we collect real world 
open-source ADS-B data from the airspace within a 50 km radius centered on two large airports (IATA: BOS, and IATA: JFK), as well as two dense airspaces above Europe and 
North America (with a radius of 100 km), and create datasets that meet the requirements of our task. Finally, the datasets we constructed comprise a 
total of 45,463 trajectory records. Each record includes five key attributes: timestamp, aircraft identification number, longitude, latitude, 
and altitude. Given the inconsistency in the formatting of aircraft identification numbers, we apply encoding to standardize these identifiers, 
thereby facilitating more effective model training. The time interval between consecutive trajectory records for each aircraft is 10 seconds, 
based on the ADS-B data we collected.

We quantitatively evaluate the model's performance using the Average Displacement Error (ADE) and Final Displacement Error (FDE), 
which are commonly employed in trajectory prediction tasks \cite{mohamed2020social}. The specific mathematical expressions are provided in Eq. (13) and Eq. (14).
\begin{equation}
\label{equation13}
A D E=\frac{\sum_{i \in N} \sum_{t \in T_{p}}\left\|\hat{p}_{t}^{i}-p_{t}^{i}\right\|_{2}}{N \times T_{p}}
\end{equation}
\begin{equation}
\label{equation14}
F D E=\frac{\sum_{i \in N}\left\|\hat{p}_{t}^{i}-p_{t}^{i}\right\|_{2}}{N}
\end{equation}
where $p_t^i$ and ${\hat{p}}_t^i$ represent the actual and predicted horizontal or vertical trajectories of the aircraft, respectively. 
$T_p$ denotes the prediction time horizon, and $N$ represents the number of aircraft. Additionally, when evaluating the model's performance, 
we calculate the ADE and PDE for both latitude/longitude and altitude separately, as their units are not uniform.

\subsection{Experiment settings}
Table 1 presents our experimental parameter settings. The DA-STGCN model is trained with a batch size of 128, using the ReLU 
activation function (LeakyReLU is employed in the GAT). We utilize the Adam optimizer with an initial learning rate of 0.001, 
which decays to 0.0002 after 200 epochs, with a total of 400 epochs trained. In the experimental process, the model predicts the trajectory distribution 
for the next 60 seconds based on the observed states from the preceding 40 seconds. The implementation is based on PyTorch and trained on four NVIDIA RTX 2080 Ti GPUs.
\begin{table}[t]\small
\centering
\begin{tabular}{p{5.6cm}p{1.3cm}}
\toprule
Model hyperparameters     & Value     \\ \midrule
Batch size                & 128       \\
Observation time step     & 40 s      \\
Prediction time step      & 60 s      \\
STGCN layer number        & (1,3,5,7) \\
TXP-CNN layer number       & (1,3,5,7) \\
Attention head number     & (4,1)     \\
Optimizer                 & Adam      \\
Number of training epochs & 400       \\
Learning rate             & 0.001     \\ \bottomrule
\end{tabular}
\caption{DA-STGCN model parameter settings.}
\label{table1}
\end{table}
\subsection{Analysis of experimental results}
\textbf{Selection of network layers.} We first conduct an ablation study on the number of layers for STGCN and TXP-CNN, comparing their predictive performance 
with 1, 3, 5, and 7 layers, respectively. The experimental results are presented in Table 2, and it indicate that the DA-STGCN 
model achieves the best overall performance in aircraft trajectory prediction when the number of layers for STGCN and TXP-CNN is set to 1 and 5, respectively. 
Therefore, all subsequent experiments are based on this layer configuration.
\begin{table}[t]\small
\centering
\setlength{\tabcolsep}{1mm}
\begin{tabular}{ccccc}
\toprule
\makecell{Layer \\number}    & 1            & 3            & 5            & 7            \\ \midrule
1 & 0.0079/33.1 & 0.0082/37.6 & \textbf{0.0077/31.7} & 0.0082/34.7 \\
3 & 0.0074/35.6 & 0.0091/39.3 & 0.0079/33.8 & 0.0120/41.2  \\
5 & 0.0100/40.1  & 0.0093/52.8 & 0.0075/62.2 & 0.0076/36.2 \\
7 & 0.0093/32.5 & 0.0086/32.9 & 0.0110/30.4  & 0.0092/30.2 \\ \bottomrule
\end{tabular}

\caption{Ablation Study on the Selection of Layers for STGCN and TXP-CNN. The indices listed in first column and the first row represent 
the layer numbers of STGCN and TXPCNN respectively. In the table, the left numbers correspond to the ADE for the aircraft's horizontal trajectory, 
and the right numbers correspond to the ADE for the vertical trajectory.}
\label{table2}
\end{table}

\begin{table*}[t]
\centering
\begin{tabular}{@{}ccccccc@{}}
\toprule
\multirow{2}{*}{Models} & \multicolumn{2}{c}{Airport Terminal Area$\downarrow$}     & \multicolumn{2}{c}{Dense Airspace$\downarrow$}           & \multicolumn{2}{c}{Average$\downarrow$}                  \\ \cmidrule(l){2-7} 
                        & Horizontal           & Vertical             & Horizontal            & Vertical             & Horizontal            & Vertical             \\ \midrule
LSTM                   & 0.014/0.022            & 55.80/60.46          & 0.025/0.027           & 77.46/80.85          & 0.020/0.025           & 66.63/70.66          \\
S-LSTM                 & 0.013/0.020            & 54.37/55.59          & 0.023/0.024           & 75.19/78.84          & 0.018/0.022           & 64.78/67.22          \\
STGAT                 & 0.012/0.019            & 49.99/50.44          & 0.015/0.017           & 67.51/75.80          & 0.014/0.018           & 58.78/63.12          \\
Social-STGCNN               & 0.0096/0.013           & 42.54/48.79          & 0.012/0.016           & 63.12/75.54          & 0.011/0.015           & 52.83/62.17          \\
w/o GAT(Ours)                & 0.0083/0.011           & 36.10 /36.20         & 0.010/0.012           & 50.37/54.08          & 0.092/0.012           & 43.24/45.14          \\
w/o Attention Module(Ours) & 0.0082/0.011           & 34.31/40.80          & 0.010/0.014           & 52.43/61.01          & 0.0091/0.013          & 43.37/50.91          \\
\textbf{DA-STGCN}(Ours)      & \textbf{0.0068/0.0093} & \textbf{27.47/32.35} & \textbf{0.0095/0.012} & \textbf{48.80/57.16} & \textbf{0.0082/0.011} & \textbf{38.14/44.76} \\ \bottomrule
\end{tabular}
\caption{Performance of Different Prediction Models. 
Within the table, the terms ``Horizontal'' and ``Vertical'' correspond to the model's predictive performance in their respective dimensions. 
The left-hand side of the table displays the ADE values, illustrating the average prediction accuracy, while the right-hand side presents the FDE values, 
highlighting the precision of the model's endpoint predictions. $\downarrow$ indicates that the smaller these two metrics, the better the performance of the model.}
\label{table3}
\end{table*}

\textbf{Quantitative analysis of model performance.} We test and compare the predictive performance of the DA-STGCN model on two datasets: 
the airport terminal area and the dense airspace. We select widely used and high-performance trajectory prediction models in the current 4D trajectory prediction 
domain as baseline methods for the comparative experiments. Table 3 provides a detailed presentation of the results. 
Overall, DA-STGCN outperforms existing methods on both metrics. Compared to Social-STGCNN, our model achieves approximately a 26\% reduction in average ADE 
and a 27\% reduction in FDE. This performance difference is due to the fact that Social STGCNN only uses the adjacency matrix pre-defined by the kernel function 
to describe the relationships between nodes, which cannot capture the global features between nodes well. 
In the airport terminal area, the interactions between aircraft are more significant, as many aircraft have not yet entered a steady flight state or are in the process of 
landing. Our proposed Dual-Attention module can extract more valuable information in such scenarios, which benefits the model's predictive performance.
    
To verify the effectiveness of our proposed Dual-Attention module, we design corresponding ablation experiments. The results are shown in the last three rows of Table 3. 
The results demonstrate that the Dual-Attention module effectively improves model performance, with the best performance achieved when 
both attention modules are used. We attribute this to the fact that the two attention modules focus on different features, providing complementary information.

\textbf{Qualitative analysis of model performance.} The quantitative analysis section demonstrates that the DA-STGCN model outperforms existing 4D trajectory 
prediction methods on both metrics. In this section, we conduct a qualitative analysis of the trajectory prediction results. Figure 5 illustrates the 
visualization of predicted trajectories from the DA-STGCN and Social-STGCNN models in the same scenario. 
It is worth noting that the aircraft 5 in Figure 5(a) and Figure 5(c) 
have not yet taken off, so their trajectory distributions are not present. As shown in the results of Figure 5, the predicted trajectories from the 
DA-STGCN model closely reflect the aircraft's actual future trajectories and outperform those from the Social-STGCNN model. In the DA-STGCN model's predictions, 
the trajectories of aircraft 1 and 2 are nearly identical to their actual future trajectories, whereas the Social-STGCNN model exhibits significant deviations 
between the predicted and actual trajectories for these two aircraft, because it misjudged the mutual influence between these two aircraft. 
This indicates that the DA-STGCN model more effectively leverages the potential correlations 
between nearby aircraft, resulting in more accurate predictions. For altitude predictions, the DA-STGCN model effectively predicts each aircraft's future height, 
despite some minor deviations. However, the Social-STGCNN model shows significant errors in altitude predictions for aircraft 3, which is undesirable. Overall, 
the DA-STGCN model outperforms the Social-STGCNN model in predicting future aircraft trajectories, demonstrating the effectiveness of our method.
\begin{figure}[t]
\centering
\includegraphics[width=1\columnwidth]{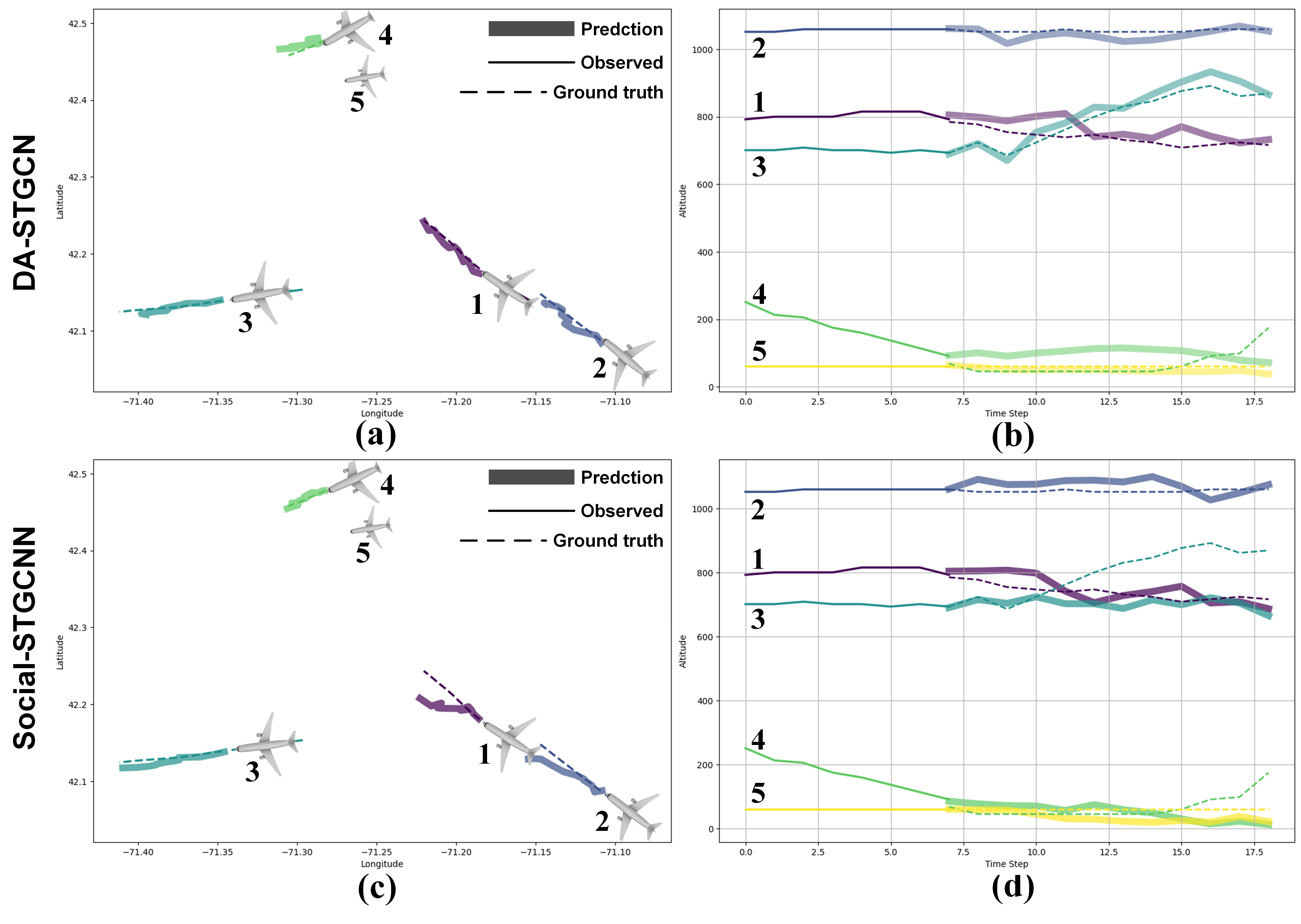} 
\caption{Qualitative Analysis of Aircraft Trajectory Prediction Results. (a) and (b) show the trajectory distributions in the horizontal and vertical dimensions predicted by the DA-STGCN model for a 
scenario in the airport terminal area, while (c) and (d) display the corresponding predictions by the Social-STGCNN model.}
\label{fig5}
\end{figure}

\begin{figure}[t]
\centering
\includegraphics[width=0.95\columnwidth]{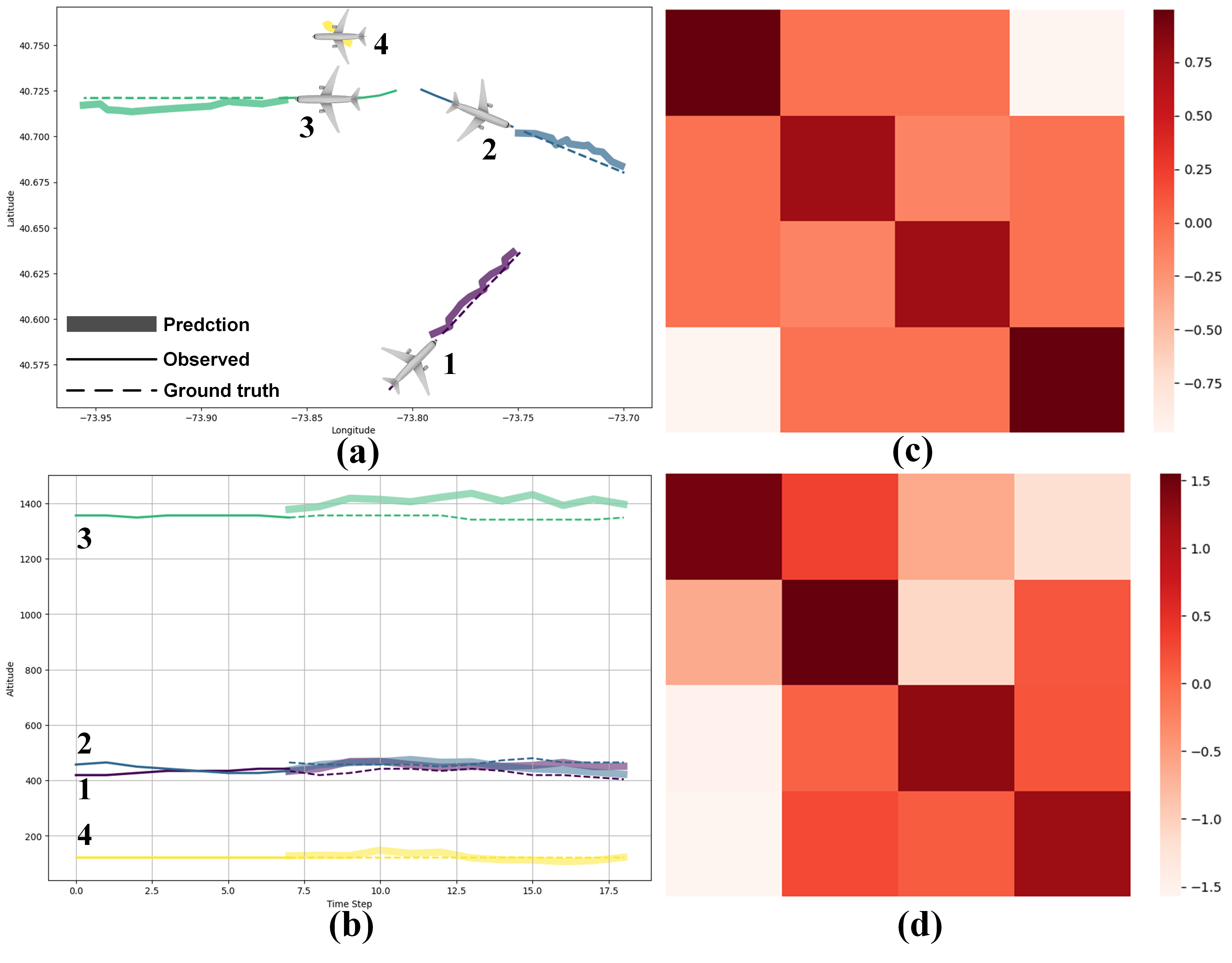} 
\caption{Qualitative Analysis of Self-Attention Reconstructed Adjacency Matrix. (a) and (b) display the trajectory distributions in the horizontal 
and vertical dimensions, respectively, while (c) and (d) show the initial and reconstructed adjacency matrices. 
The value in a specific row and column of the adjacency matrix represents the extent of influence that the aircraft in that row has on the aircraft in that column.}
\label{fig6}
\end{figure}

We proceed to undertake a qualitative examination of the adjacency matrix that has been reconstructed utilizing the self-attention mechanism, 
offering insights into the relational structure captured by the model.
Figure 6 shows the aircraft trajectory prediction results and the corresponding adjacency matrices from the DA-STGCN model for a scenario in the 
airport terminal area. 
A direct comparison of Figures 6(c) and 6(d) reveals 
that the reconstructed adjacency matrix contains more comprehensive feature information, with more pronounced interactions between aircraft. In Figure 6(d), 
the influence between aircraft 1 and aircraft 3 is less than in Figure 6(c). This is because these two aircraft are flying in completely opposite directions 
and have a significant altitude difference. Our self-attention reconstruction module captures this information and optimizes the adjacency matrix, 
which is highly meaningful. Additionally, the initial adjacency matrix suggests a strong correlation between aircraft 2 and aircraft 3 due to their 
close proximity, which is inaccurate. Aircraft 2 and aircraft 3 are flying in opposite directions, and their influence on each other is not as significant as 
their distance might suggest. Consequently, the reconstructed adjacency matrix reduces the value in this case, as seen in the entry in the second 
row and third column. Overall, our self-attention reconstruction module effectively optimizes the adjacency matrix, capturing more accurate and 
comprehensive node-related information.
\section{Conclusion}
In this paper, we propose a DA-STGCN model to address the limitations of current aircraft trajectory prediction methods, which often fail to effectively capture 
the potential correlations between aircraft. Our model first reconstructs the adjacency matrix using self-attention mechanisms to capture more accurate and 
comprehensive node-related information, rather than relying solely on distance. Subsequently, we employ GAT for additional node feature aggregation, enriching 
the initial embeddings and significantly bolstering the model's capacity to discern underlying node correlations.
Our model has been validated on two real ADS-B datasets, and the quantitative analysis of the experimental results demonstrates that the 
DA-STGCN model shows superior performance over existing methods in terms of both ADE and FDE metrics. The qualitative analysis indicates that our proposed 
dual-attention module effectively optimizes the adjacency matrix's representation of node information and aggregates more comprehensive additional data. 
This not only demonstrates the superiority of our approach but also highlights its potential to drive significant progress in the field of multi-object trajectory prediction.




\section{Acknowledgments}
This work is supported by the National Natural Science Foundation of China (No. U2333211), and grants from Science \& Technology Department of Sichuan Province of China (Nos. 24JBGS0075, 2024YFHZ0095, 2023YFG0054).


\bibliography{aaai25}

\end{document}